# The Elusive Quest for Intelligence in Artificial Intelligence

Dr Shoumen Palit Austin Datta
MIT Auto-ID Labs
Massachusetts Institute of Technology

The elusive quest for intelligence in artificial intelligence prompts us to consider that instituting human-level intelligence in systems may be (still) in the realm of utopia. In about a quarter century, we have witnessed the winter of AI (1990) being transformed and transported to the zenith of tabloid fodder about AI (2015). The discussion at hand is about the elements that constitute the canonical idea of intelligence. The delivery of intelligence as a pay-per-use-service, popping out of an app or from a shrink-wrapped software defined point solution, is in contrast to the bio-inspired view of intelligence as an outcome, perhaps formed from a tapestry of events, cross-pollinated by instances, each with its own microcosm of experiences and learnings, which may not be discrete all-or-none functions but continuous, over space and time. The enterprise world may not require, aspire or desire such an engaged solution to improve its services for enabling digital transformation through the deployment of digital twins, for example. One might ask whether the "work-flow on steroids" version of decision support may suffice for intelligence? Are we harking back to the era of rule based expert systems? The image conjured by the publicity machines offers deep solutions with human-level AI and preposterous claims about capturing the "brain in a box" by 2020. Even emulating insects may be difficult in terms of real progress. Perhaps we can try to focus on worms (*Caenorhabditis elegans*) which may be better suited for what business needs to quench its thirst for so-called intelligence in AI.

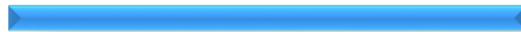

# The Elusive Quest for Intelligence in Artificial Intelligence

Is intelligence an illusion in AI? This essay is an academic reflection. The opinions are solely due to the author.

Dr Shoumen Palit Austin Datta, MIT Auto-ID Labs, Massachusetts Institute of Technology, Cambridge, Massachusetts, USA

**PLEASE OPEN DOCUMENT WITH ADOBE ACROBAT TO ACTIVATE THE EMBEDDED HYPERLINKS**

The promise of and the pessimism about AI (artificial intelligence) charts a sinusoidal path. The questions about "intelligence" in AI persists. Even more questions failed to answer if the I in AI is true intelligence or some advanced form of multi-level dynamics with/without a higher order of structured complex networks or (even better) temporal networks. The semantics depends on and is colored by one's view of systems or interpretation of what is intelligence, what constitutes proof of intelligence and the nature of complexity.

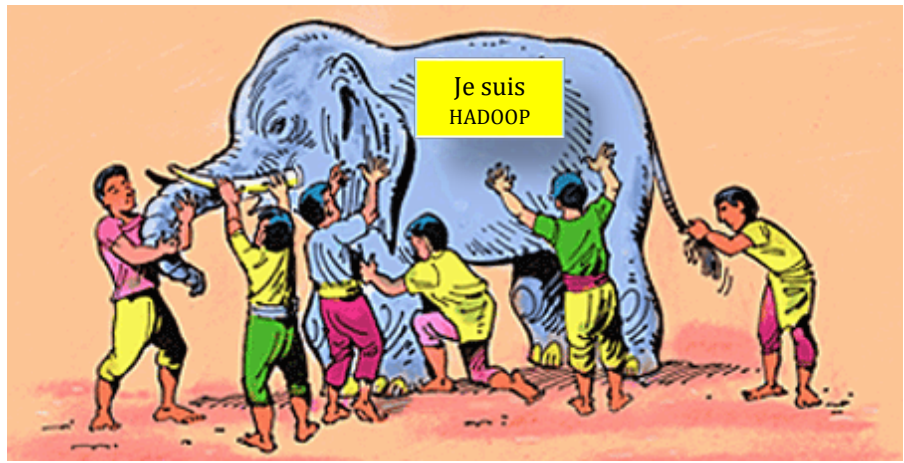

The Blind Men and the Elephant ● John Godfrey Saxe (1816-1887)

In systems thinking a common analogy is that of six blind men from Indostan (India) touching various parts of an elephant and claiming that "elephant is a tree" (man who touched the leg), "it is like a rope" said the blind man who touched the tail. "Like a snake" (man who touched the trunk). "It is like a big hand fan" (man who touched the ear). "Like a huge wall" said the man who touched the belly of the elephant. "Like a spear" said the blind man who touched the tusk. They were misguided in describing their part and wrong about the whole picture. The definition of intelligence may be colored by the professional bias of the interpreter. Intelligence is not a point. It is a line. It is an array or collective continuum of network systems which may not be boxed with human skills, yet (the latter is by far the most ludicrous and incredible claim at the present time under our known circumstances).

### *Why this mad pursuit?*

University of Cambridge which had done best at teaching mathematics is the one from amongst whose graduates have come more of the English poets, while Oxford which has specialized in the humanities, has tended to turn out writers who have attained, on the whole, a high level of mediocrity. I suppose that by the time one has discussed literature with a witty and learned professor, you know what has been achieved and how good it is. You become respectful and begin to wonder who am I to do better? (Dialogues of Alfred North Whitehead (1861-1947) as recorded by Lucien Price. Little, Brown ▪ Boston, 1954)

The [definitive treatise](#) by Stuart J Russell and Peter Norvig (*Artificial Intelligence: A Modern Approach*, 3rd edition, 2015) should suffice to discourage all amateurs (myself included) to desist from testifying about the trials and tribulations with respect to I in AI.

### *Who am I to do better?*

In this [neo-Norvig-ean](#) era of human-level AI we find [LISP](#) functional programming language, the 2nd oldest language, since FORTRAN. It is still regarded as a powerful AI tool. Figure 1 in the first LISP paper ([John McCarthy, MIT](#)) illustrates "Representation of S-Expressions by List Structure" which takes us back to the [history of perceptrons](#) elaborated in the [seminal paper](#) by Warren McCulloch and Walter Pitts (1943) *A Logical Calculus of the Ideas Immanent in Nervous Activity* in Bulletin of Mathematical Biophysics **5** 115-133 (marks the dawn of AI).

The [perceptron algorithm](#) created by Frank Rosenblatt was championed by the US Navy in fueling the "intelligence" controversy to the extent that *The New York Times* reported (in 1958) the perceptron to be "the embryo of an electronic computer that [the Navy] expects will be able to walk, talk, see, write, reproduce itself and be conscious of its existence." A decade later, Marvin Minsky and Seymour Papert observed (in [Perceptrons, MIT, 1969](#)) that such statements may be guilty of the wildest exaggeration. (However, Minsky is no stranger to extending the euphoria and feeding the frenzy with [predictions](#) about AI's rosy potential. Extending the scope of intelligence in AI has always been in vogue and [continues](#) unabated.)

It is well nigh impossible to over emphasize that perceptrons, decision trees, recursion functions, etc, are good topological representations and manifestations of the anatomy which are visible in the context of neural connectivity and the neural networks which we observe, for example, in worms, eg *Caenorhabditis elegans* ([Robert Horvitz, MIT; figure 3](#)).

Abstractions of these networks, eg, cube on cube ([The Society of Mind](#) by Marvin Minsky, MIT, 1985) and "learning" mechanisms ([The Organization of Behavior](#) by Donald O Hebb, 1949) may not constitute intelligence. These processes populate fields with values from the user environment which can be selectively used (*per contra* hard coded defined sets). For example, [NEST](#) Learning Thermostat uses input values to tune *your preferred* temperatures.

Elements of the equation/rule -based (brittle and static) structures caused the bust of the expert systems and ended *The AI Business* (Winston and Prendergast, MIT, 1984) lure before the rise of artificial neural networks (ANN popularity circa 1990). Topology and synaptic weights, if combined, offered a flexible infrastructure to acquire more relevant values and profit from data. This is an important advancement. But, is it really intelligence?

Rather than partial differential equations exploding due to increase in state functions (due to the large number of parameters), Agents allowed each variable to be represented as a single-function entity. The collective output from an Agency of Agents improved predictive or prescriptive precision compared to operations research applications (see illustration below). The behavior of Agents and Agencies using "AI" concepts originated from the foundation laid by the principles of stigmergy (Pierre-Paul Grasse, 1959) which continues to evolve as artificial life and are able to address complex business problems.

The recent surge in the hype associated with "big data" has navigated profitability from analytics to the front and center. Intelligence is marketed as a commodity in this scenario.

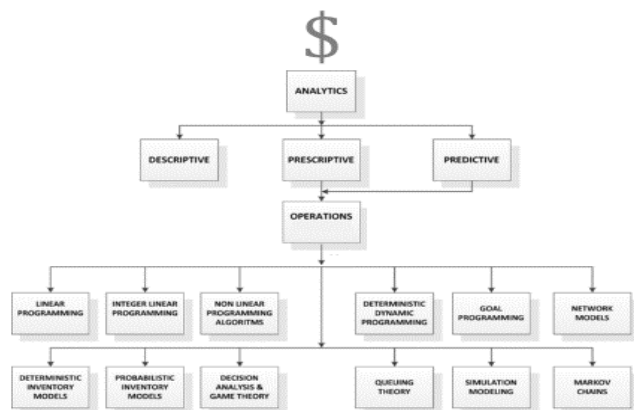

In order to market intelligence as a service, the AI paradigm is being refurbished as a commodity (brain in a box) and touted to business and industry as an essential tool to reach the luminous summit [$]. Winning at games using ANN is advocated as intelligence. Smart and intelligence are emerging as speculative tabloid fodder. Witnessing the rapid transmutation of tabloid fodder about speculation to (business) truth is deeply troubling.

Claims about "original" thinking in containerization of data and process are good ideas but appeared as concepts proposed almost half a century ago by Marvin Minsky (page 315 in the original book or search page 311 in this PDF version of the book). Connecting entities (containers) using IPv6 resonates with ideas suggested about a decade ago. However, it is reassuring that the concepts are not lost but are being developed to advance the march of digital transformation (see Digital Twins https://dspace.mit.edu/handle/1721.1/104429).

### *Is Learning a Myth in "Human-level" AI Systems?*

I cannot improve the content from [Rodney Brooks](#), hence, I shall quote verbatim as follows: "We have found a way to build fixed topology networks of our finite state machines which can perform learning, as an isolated subsystem, at levels comparable to these examples. At the moment of course we are in the very position we lambasted most AI workers for earlier in this paper. We have an isolated module of a system working and the inputs and outputs have been left dangling." ([Intelligence without Representation, 1991](#))

Learning triggers profound, sustained, often long term changes in our neural networks at many levels that we cannot even begin to understand or grasp its cognitive repercussions. Thus, almost all assumptions made by [McCulloch & Pitts (1943)](#) are violated (Appendix 1).

The "all or none" phenomena assumed by [McCulloch & Pitts (1943)](#) is only relevant from a mechanical perspective if one assumes (incorrectly, of course) that input data is supposed to transduce a signal and the resultant action potential (neuronal activation) may be one form of a proof of learning. Neurologists will strenuously and vociferously take exception.

AI experts may wish adopt this view to claim learning in the AI context. The neurological state of learning, cognition and behavior is usually a continuous function modulated by evolutionary weights, which are not subject, in the least, to the limitations of discrete-state machines. Application of machine learning models are often [inconsistent and incorrect](#).

Discrete systems have a finite (countable) number of states which may be described in precise mathematical models. The computer is a finite state machine which may be viewed as a discrete system. The brain is not a computer. The neural infrastructure and networks are not finite state machines. Imposing any such model (real-world continuous systems) or ill-advised abstraction or gross extrapolation (by those not so well informed) may only perpetuate great lengths of fantasy about intelligence and learning related to AI systems.

"Of the vast stream of sense data that pour into our nervous systems we are aware of few and we name still fewer. For it is the fact that even percepta are wordless. Only by necessity do we put a vocabulary to what we touch, see, taste, and smell, and to such sounds as we hear that are not themselves words. We look at a landscape, at the rich carving and majestic architecture of a cathedral, listen to the development of harmonies in a symphony, or admire special skill in games and find ourselves woefully lacking in ability to describe our percepts. Words, as we very rightly say, fail us either to describe the plain facts of these experiences or to impart to others, our feelings." (G Jefferson CBE, FRS, MS, FRCS, Professor of Neurosurgery in *[The Mind of Mechanical Man](#)* in British Medical Journal, 25th June 1949). The [author was aware](#) of "Dr Wiener of Boston, his entertaining book *[Cybernetics](#)* (1948)."

Alan Turing was [cognizant](#) of the over-reach in claiming "intelligence" in AI and outlined potential objections including Godel's theorem ([mathematical objection](#)) and "Argument from Consciousness" which he reproduced from [Professor Geoffrey Jefferson](#) as a quote (from his Lister Oration, 1949) "Not until a machine can write a sonnet or compose a concerto because of thoughts and emotions felt, and not by the chance fall of symbols, could we agree that machine equals brain - that is not only write it but know that it had written it. No mechanism could feel (and not merely artificially signal, an easy contrivance) pleasure at its successes, grief when its valves fuse, be warmed by flattery, be made miserable by its mistakes, be charmed by sex, be angry or depressed when it cannot get what it wants."
A. M. Turing (1950) Computing Machinery and Intelligence. Mind **49** 433-460 ([PDF](#))

Page 452 (see Appendix 2) removes any doubt that Turing had grave doubts regarding claims of intelligence in the context of computers. Turing's suggested starting point is "the child machine" (Appendix 2). Then he proposes to add the roles or processes of "evolution" "hereditary material" "mutation" "education" and "natural selection" in order to mature "the child machine" to "imitate an adult human mind" as a path forward to intelligence. To understand even vaguely what happens after "the initial state of mind, say at birth" the reader is urged to review [Patterns in the Mind](#) by [Ray Jackendoff](#) (1966) and then take into consideration the field of [linguistics](#) and [natural language](#) development (1970, PhD thesis of Terry Winograd, MIT [http://hci.stanford.edu/winograd/shrdlu/AITR-235.pdf](http://hci.stanford.edu/winograd/shrdlu/AITR-235.pdf)).

For all this to happen, we must process information encoded via developmental and environmental signals. Hence, the suggestion, research and convergence *on* the concept of molecular logic gates. The [complexity of the process](#) may help deter one from concluding that we *are* dealing with intelligence with respect to computers, machinery or AI systems.

However, the human spirit and the fabric of scientific research cannot step away from problems even if all available reason suggests that something is impossible, at the time. It is with this fervor the 1956 Dartmouth Summer Research Project on Artificial Intelligence (June 17 - August 16) was [proposed](#) in 1955 by a visionary group of eminent and erudite academic scholars ([www.aaai.org/ojs/index.php/aimagazine/article/view/1904/1802](http://www.aaai.org/ojs/index.php/aimagazine/article/view/1904/1802)).

The [proposal](#) (see Appendix 3) admits it is a "conjecture" but continued "that every aspect of learning or any other feature of intelligence can in principle be so precisely described that a machine can be made to simulate it."

Great strides (Appendix 4) have been made, yet the 1956 Summer Research "conjecture" looms overhead. But, our "faith" in progress of AI is evident from the 1145 page book by Russell and Norvig ([AI - A Modern Approach, 3rd ed](#)). We are learning how [decisions can be made without a brain](#) in cognitive organisms (unicellular mould *Physarum polycephalum*).

### *Neurobiology 101 - neurons - their numbers and networks*

Topology and weights are the foundational underpinnings of artificial neural nets (ANN) which are the mainstay of AI systems. How reliable are these extrapolations? Are we still talking about AI? Let me reiterate what [Rodney Brooks has stated](#) but in a different vein.

Structural design of network topology aims to mimic the commonly observed *organization* of neurons. Topology based on neural organization ([small world networks](#)) may be fraught with errors as evidenced by studies on [wiring configuration](#) and [neuroanatomical analysis](#) which reveals differences in [circuit architecture](#) and connectivity if viewed at [mesoscopic](#) vs microscopic scale. On a mesoscopic scale, seemingly random networks exhibit consistent properties. It may be difficult, if not impossible, to extract useful/meaningful abstractions from these counter-intuitive [non-linear yet dynamic](#) structure-function complementarities.

In ANN, weights are assigned to signify connectivity strengths (the links between the perceptrons). These are arbitrary, at best, because synaptic weights between neurons and clusters are subjected to conditions that we think we know only by name. Even if one acquired neurophysiological data related to frequency variations of action potentials (~200 Hertz) in an attempt to understand synaptic weights between neurons, the results may not be revealing. The complexity may be compounded by the fact nerve transmissions are modified by ions, electrical threshold potential and chemical neuro-transmitters.

In synaptic design, one assumes the all-or-none process ([Appendix 1](#)) and the weights are modeled based on extrapolation from "inferential changes" which are in the order of milliseconds to seconds (hence, subject to observation, data collection and extrapolation). But, the nature of the connectivity and resultant weight is also influenced by epigenetic factors (time scale – seconds to days), ontogenic factors (days to years) and phylogenetic factors which are the result of generations or are derived from the evolutionary time scale, as noted in [Appendix 2](#). Hence, the nature of the weight deduced from "inferential" changes (primarily sense and response mechanisms) are only the tip of the iceberg. We are almost completely in the dark about the nature of the influence from these other three factors.

Taken together, perhaps we are starting out on the wrong foot about the design of ANN by holding on to assumptions which are generally incorrect because we remain significantly uninformed. Having said that, one must hasten to add, that, no matter how approximately correct the synthetic weights, may be, it may not be impossible to conceive building ANNs with partially unqualified numbers and unsure topologies. Using tools eg back propagation algorithms, the AI system may be tuned and re-tuned in a dynamic data-driven manner (some may still refer to it as "[learning](#)") to yield actionable information from [higher order](#) systems. Over-fitting the model may cause harm, for example, collision avoidance systems.

Scholars continue to discuss new ways of using robots to make robots, create self-healing intelligent machines and adaptive machines to optimize up-time. Thinkers are conjuring up ways to harness the developmental foundations of neurons – *neurogenesis*. Emulation of *neural development* using computational AI systems can incorporate characteristics of natural neural systems into engineering design. Scientists are claiming that rather than *designing* neural networks, emulation of neurogenesis shall enable us to *generate* neural networks to serve dynamic and even more complex systems of the future.

This emerging field of programmable artificial neurogenesis appears to call for a meta-design paradigm which may begin with components (object oriented?). It aims to build higher order intelligent systems which will adapt (to demands, environment, resource) without re-programming component level entities. When components are updated, the changes will be propagated, via appropriate "learning" functions, up/down hierarchies.

The great desire to emulate the grand vision latent in intelligence, cognition and the brain, works almost as an aphrodisiac. The immense powers of biology and the ability to distil and capture even an iota of that potential in bio-inspired systems through convergence with computation will continue to be a Holy Grail. Here is one example of bio-power:

We have about 3 billion base pairs (A-T, G-C) in the human genome ($3 \times 10^9$) which codes for about 10,000 – 20,000 genes resulting in a human body with 100 trillion cells ($1 \times 10^{14}$).

At least a third of the approximately 20,000 different genes that make up the human genome are active (expressed) in the brain. We have about $8.5 \times 10^{10}$ neural cells (there are an equivalent number of glial cells). Each neural cell connects on an average with 1,000 other neural cells to create about $1 \times 10^{14}$ neural connections. This is the neural network which makes us human, creates intelligence and cognition. If we may think in terms of a compression ratio, the ratio approaches $10^{11}$ (7,000 genes creating $1 \times 10^{14}$ connections).

The most effective compression algorithm CMIX doesn't even come close. The illegal 42.zip bomb which unfolds to 4.5 petabytes (pb) from a 42 kilobytes (kb) single symbol zip approaches a compression ratio of $10^{11}$ but in an artificial circumstance devoid of any intelligence. The human compression of $10^{11}$ offers sustainable, real, life-long intelligence.

**Conclusion –** *These 'intelligent' machines may never be intelligent in a human sense (p339)*

A quantum leap, still cryptic within the unknown unknowns, may unleash intelligence in AI, in the future. We must continue to explore far and wide, emulate insects and think about the Octopus. "We can only see a short distance ahead, but we can see plenty there that needs to be done." Convergence of tools (statistics, math) with data curation (noise vs signal) is replete with promise and profitability even if it lacks (human-level) intelligence.

# APPENDIX – 1

Warren S McCulloch and Walter H Pitts (1943) *A Logical Calculus of the Ideas Immanent in Nervous Activity*. Bulletin of Mathematical Biophysics **5** 115-133
http://www.cse.chalmers.se/~coquand/AUTOMATA/mcp.pdf

### A Logical Calculus of Ideas Immanent in Nervous Activity

adequate. But for nets undergoing both alterations, we can substitute equivalent fictitious nets composed of neurons whose connections and thresholds are unaltered. But one point must be made clear: neither of us conceives the formal equivalence to be a factual explanation. *Per contra!*—we regard facilitation and extinction as dependent upon continuous changes in threshold related to electrical and chemical variables, such as after-potentials and ionic concentrations; and learning as an enduring change which can survive sleep, anaesthesia, convulsions and coma. The importance of the formal equivalence lies in this: that the alterations actually underlying facilitation, extinction and learning in no way affect the conclusions which follow from the formal treatment of the activity of nervous nets, and the relations of the corresponding propositions remain those of the logic of propositions.

The nervous system contains many circular paths, whose activity so regenerates the excitation of any participant neuron that reference to time past becomes indefinite, although it still implies that afferent activity has realized one of a certain class of configurations over time. Precise specification of these implications by means of recursive functions, and determination of those that can be embodied in the activity of nervous nets, completes the theory.

### THE THEORY: NETS WITHOUT CIRCLES

We shall make the following physical assumptions for our calculus.

1. The activity of the neuron is an "all-or-none" process.

2. A certain fixed number of synapses must be excited within the period of latent addition in order to excite a neuron at any time, and this number is independent of previous activity and position on the neuron.

3. The only significant delay within the nervous system is synaptic delay.

4. The activity of any inhibitory synapse absolutely prevents excitation of the neuron at that time.

5. The structure of the net does not change with time.

# APPENDIX – 2

Page 452

A. M. Turing (1950) Computing Machinery and Intelligence. Mind **49** 433-460
www.csee.umbc.edu/courses/471/papers/turing.pdf

> In the process of trying to imitate an adult human mind we are bound to think a good deal about the process which has brought it to the state that it is in. We may notice three components.
>
> (a) The initial state of the mind, say at birth,
>
> (b) The education to which it has been subjected,
>
> (c) Other experience, not to be described as education, to which it has been subjected.
>
> Instead of trying to produce a programme to simulate the adult mind, why not rather try to produce one which simulates the child's? If this were then subjected to an appropriate course of education one would obtain the adult brain. Presumably the child brain is something like a notebook as one buys it from the stationer's. Rather little mechanism, and lots of blank sheets. (Mechanism and writing are from our point of view almost synonymous.) Our hope is that there is so little mechanism in the child brain that something like it can be easily programmed. The amount of work in the education we can assume, as a first approximation, to be much the same as for the human child.
>
> We have thus divided our problem into two parts. The child programme and the education process. These two remain very closely connected. We cannot expect to find a good child machine at the first attempt. One must experiment with teaching one such machine and see how well it learns. One can then try another and see if it is better or worse. There is an obvious connection between this process and evolution, by the identifications
>
> Structure of the child machine = hereditary material
>
> Changes of the child machine = mutation,
>
> Natural selection = judgment of the experimenter

## APPENDIX – 3

1955 Dartmouth Summer Research Proposal
http://www.aaai.org/ojs/index.php/aimagazine/article/view/1904/1802

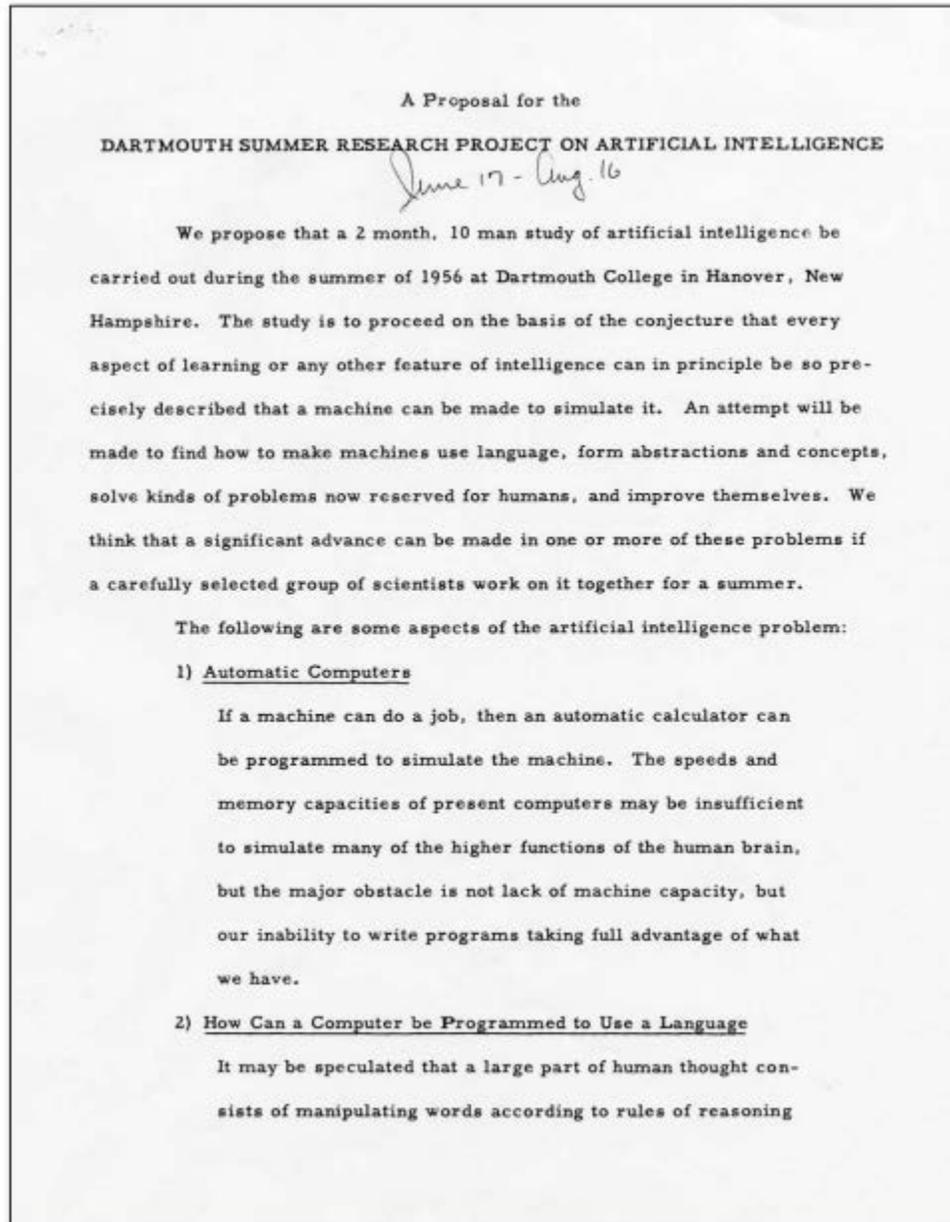

Page 1 of the Original Proposal.

## APPENDIX – 4

**1943** Warren S McCulloch and Walter H Pitts *A Logical Calculus of the Ideas Immanent in Nervous Activity* in Bulletin of Mathematical Biophysics **5** 115-133
http://www.cse.chalmers.se/~coquand/AUTOMATA/mcp.pdf

**1949** Geoffrey Jefferson *The Mind of Mechanical Man* in British Medical Journal June 25
http://www.ncbi.nlm.nih.gov/pmc/articles/PMC2050428/pdf/brmedj03683-0003.pdf

**1950** Alan M. Turing *Computing Machinery and Intelligence* in Mind 49 433-460
www.csee.umbc.edu/courses/471/papers/turing.pdf

**1955** Proposal for Dartmouth Summer Research Project on Artificial Intelligence
http://www.aaai.org/ojs/index.php/aimagazine/article/view/1904/1802

**1960** Marvin Minsky *Steps Toward Artificial Intelligence* in Proceedings of the IRE
http://web.media.mit.edu/~minsky/papers/steps.html
http://www.cs.utexas.edu/~jsinapov/teaching/cs378/readings/W2/Minsky60steps.pdf

**1966** Ray Jackendoff *Patterns in the Mind*
http://ling.umd.edu/~omaki/teaching/Ling240_Summ2007/Jackendoff94_Ch1.pdf

**1970** Natural Language – PhD Thesis of Terry Winograd, MIT
http://hci.stanford.edu/winograd/shrdlu/AITR-235.pdf

**1974** Marvin Minsky *A Framework for Representing Knowledge* in MIT AI Lab Memo 306
www.u-picardie.fr/~furst/docs/Minsky_Frames_1974.pdf

**1984** Patrick H Winston *Artificial Intelligence: A Perspective*
https://mitpress.mit.edu/sites/default/files/titles/content/9780262570770_sch_0001.pdf

**1984** Winston & Prendergast *The AI Business* https://mitpress.mit.edu/books/ai-business

**1986** Kay Rodgers – US Library of Congress http://files.eric.ed.gov/fulltext/ED271107.pdf

**1992** P Winston *Artificial Intelligence* 3rd http://courses.csail.mit.edu/6.034f/ai3/rest.pdf

**2015** Stuart J Russell and Peter Norvig *Artificial Intelligence - A Modern Approach* (3rd ed)
www.amazon.com/Artificial-Intelligence-Approach-Russell-2015-01-01/dp/B0182Q542W

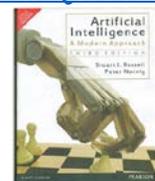

Page 11 ● I am an AI optimist. My article on AI (Agents: Where Artificial Intelligence Meets Natural Stupidity) is here http://dspace.mit.edu/handle/1721.1/41914 (2002) ▪ Dr Shoumen Palit Austin Datta, Research Affiliate, MIT Auto-ID Labs, Department of Mechanical Engineering, MIT ▪ Please explore website http://autoid.mit.edu